\PassOptionsToPackage{unicode}{hyperref}
\PassOptionsToPackage{hyphens}{url}
\PassOptionsToPackage{dvipsnames,svgnames,x11names}{xcolor}
\documentclass[
  12pt]{article}

\makeatletter
\expandafter\gdef\csname r@app:bb-example\endcsname{{A}{1}{}{}{}}
\expandafter\gdef\csname r@app:cost-transition-verification\endcsname{{D.3}{47}{}{}{}}
\expandafter\gdef\csname r@app:extensions-verifications\endcsname{{D}{37}{}{}{}}
\expandafter\gdef\csname r@app:loss-verification\endcsname{{C.1}{16}{}{}{}}
\expandafter\gdef\csname r@ass:feature-map\endcsname{{C.1}{19}{}{}{}}
\expandafter\gdef\csname r@eq:lb-holder\endcsname{{B.7}{6}{}{}{}}
\expandafter\gdef\csname r@eq:task-fast-radius\endcsname{{C.1}{16}{}{}{}}
\expandafter\gdef\csname r@eq:validation-radii\endcsname{{C.3}{17}{}{}{}}
\expandafter\gdef\csname r@fig:sim2-calls-all\endcsname{{E.1}{49}{}{}{}}
\expandafter\gdef\csname r@lem:PI-packing-testing\endcsname{{B.10}{15}{}{}{}}
\expandafter\gdef\csname r@lem:finite-stump-errors\endcsname{{D.3}{45}{}{}{}}
\expandafter\gdef\csname r@prop:unconstrained-ols-bridge\endcsname{{D.1}{44}{}{}{}}
\expandafter\gdef\csname r@sec:laptop-implementation\endcsname{{F}{49}{}{}{}}
\expandafter\gdef\csname r@sec:pr-regression-verification\endcsname{{D.1}{37}{}{}{}}
\expandafter\gdef\csname r@tab:real-laptop-services\endcsname{{F.1}{51}{}{}{}}
\makeatother

\usepackage{amsmath,amssymb}
\usepackage{mathtools,amsthm}
\usepackage{iftex}
\ifPDFTeX
  \usepackage[T1]{fontenc}
  \usepackage[utf8]{inputenc}
  \usepackage{textcomp} 
\else 
  \usepackage{unicode-math}
  \defaultfontfeatures{Scale=MatchLowercase}
  \defaultfontfeatures[\rmfamily]{Ligatures=TeX,Scale=1}
\fi
\usepackage{lmodern}
\ifPDFTeX\else
  \AtBeginDocument{%
    \let\JASAwidehat\widehat
    \renewcommand{\widehat}[1]{\JASAwidehat{#1}}}
\fi
\ifPDFTeX\else  
\fi
\IfFileExists{upquote.sty}{\usepackage{upquote}}{}
\IfFileExists{microtype.sty}{
  \usepackage[]{microtype}
  \UseMicrotypeSet[protrusion]{basicmath} 
}{}
\makeatletter
\@ifundefined{KOMAClassName}{%
  \setlength{\parindent}{2em}
  \setlength{\parskip}{0pt}
}{%
  \KOMAoptions{parskip=false}
  \setlength{\parindent}{2em}}
\makeatother
\usepackage{indentfirst}
\usepackage{xcolor}
\usepackage{booktabs,multirow,tabularx,longtable,array}
\usepackage{graphicx}
\usepackage{caption}
\usepackage{tikz}
\usetikzlibrary{arrows.meta,positioning,shapes.geometric,calc}
\usepackage{placeins,needspace,flafter}
\usepackage{algorithm}
\usepackage{algpseudocode}
\usepackage{comment}
\ifLuaTeX
  \usepackage{selnolig}
\fi
\usepackage[authoryear,round]{natbib}
\usepackage{hyperref}
\usepackage{bookmark}

\AddToHook{begindocument/end}{%
  \NewCommandCopy{\JASAoriginalref}{\ref}%
  \RenewDocumentCommand{\ref}{s m}{%
    \IfBooleanTF{#1}{\JASAoriginalref*{#2}}{%
      \ExpandArgs{e}\IfBlankTF{%
        \getrefbykeydefault{#2}{anchor}{undefined}%
        \getrefbykeydefault{#2}{url}{}}%
        {\getrefnumber{#2}}{\JASAoriginalref{#2}}}}}

\IfFileExists{xurl.sty}{\usepackage{xurl}}{}
\def\spacingset#1{\renewcommand{\baselinestretch}{#1}\small\normalsize}

\newcommand{\E}{\mathbb E}

\newcommand{\R}{\mathbb R}
\newcommand{\cA}{\mathcal A}
\newcommand{\cD}{\mathcal D}
\newcommand{\cE}{\mathcal E}
\newcommand{\cF}{\mathcal F}
\newcommand{\cG}{\mathcal G}
\newcommand{\cH}{\mathcal H}
\newcommand{\cJ}{\mathcal J}

\newcommand{\cQ}{\mathcal Q}
\newcommand{\cR}{\mathcal R}

\newcommand{\bX}{\mathbf{X}}
\newcommand{\bx}{\mathbf{x}}
\newcommand{\bF}{\mathbf{F}}

\newcommand{\bgamma}{\boldsymbol{\gamma}}

\newcommand{\bZ}{\mathbf{Z}}

\newcommand{\btheta}{\boldsymbol{\theta}}

\newcommand{\bz}{\mathbf{z}}

\newcommand{\bzero}{\mathbf{0}}
\newcommand{\bone}{\mathbf{1}}

\DeclareMathOperator*{\argmin}{arg\,min}

\newtheorem{definition}{Definition}
\newtheorem{assumption}{Assumption}

\newtheorem{proposition}{Proposition}
\newtheorem{theorem}{Theorem}
\newtheorem{corollary}{Corollary}
\newtheorem*{corollary*}{Corollary}
\newtheorem{remark}{Remark}

\date{}
\hypersetup{
  pdftitle={Prediction with Foundation Models under Deployment Budgets},
  pdfkeywords={Black-box prediction; Cost-sensitive learning; Minimax rate; Model selection; Q-learning},
  colorlinks=true,
  linkcolor=blue,
  filecolor=Maroon,
  citecolor=Blue,
  urlcolor=Blue
}
\hypersetup{pdfauthor={Zhaohui Xu, Xiaofei Xu, Doudou Zhou}}

\begin{document}
\spacingset{1}
\title{\bfseries Prediction with Foundation Models under Deployment Budgets}
\author{%
Zhaohui Xu\textsuperscript{1}, Xiaofei Xu\textsuperscript{2}, and
Doudou Zhou\textsuperscript{1}\thanks{Corresponding author: Doudou Zhou,
ddzhou@nus.edu.sg.}\\[0.35em]
{\small \textsuperscript{1} Department of Statistics and Data Science, National
University of Singapore}\\
{\small \textsuperscript{2} School of Mathematics and Statistics, Wuhan
University}%
}
\maketitle

\bigskip
\begin{abstract}
Foundation models can supply predictions for a downstream task that draw on knowledge from extensive pretraining, but they differ in task-specific accuracy
and query cost. We develop a framework that improves prediction for the task
under a budget on the average query cost per input, using training data to learn
which models to query and how to combine their outputs with features available
without querying. The first method queries the same models for every input. The second lets them vary across inputs and uses Q-learning to decide which model to
query next and when to stop. For the first method, we show that several
individually inaccurate outputs can jointly outperform the best single one. This method attains the minimax rate when the outcome depends on the outputs only through an affordable collection of models. Otherwise, every method within the budget has a slower worst-case rate, which the first method attains. For the second method, we show when varying the queried models across inputs strictly improves on querying the same models for every input, and bound its prediction risk and average cost in finite samples. In simulations and a multimodal laptop price
application, both methods outperform existing methods.
\end{abstract}

\noindent \textit{Keywords:} Black-box prediction; Cost-sensitive learning; Minimax rate; Model selection; Q-learning
\vfill

\newpage
\spacingset{1.8} 

\section{Introduction}
\label{sec:intro}

Foundation models have changed the way prediction models are built. Instead of
fitting a model to the task's own data from scratch, one can now query these
models for predictions that reflect knowledge acquired during extensive
pretraining \citep{guo2025deepseek,hollmann2025tabpfn}.
However, using these predictions alone is not enough, because the models were not
trained for the task and the task's data contain information that the models
never see. It is more sensible to use these data to adjust the output of a
foundation model and combine it with the features of the task. Moreover, several
models usually carry more information than one, since the inputs often contain
several modalities, different models read different parts of them, and models
trained on different data make different errors. Their outputs can therefore
complement one another, but they can also overlap with one another or with the
features. Whether a model is worth using thus depends on how much information it adds beyond the other models and the
features. How to combine the outputs of several foundation models with the
features is a new statistical problem.

Each additional model, however, comes at a cost. Commercial models charge for every query and locally run models consume computing
time, so the cost grows with the number of inputs to be predicted. The Amazon laptop
listings analyzed in Section~\ref{sec:real-data}, for example, each come with
numerical specifications, a text description and a product image. A tabular
foundation model can use the specifications, language models the description and
vision-language models the image, but their prices per query differ by nearly
three orders of magnitude. With a budget, one must also decide which of these models to query before combining their outputs.
Whether a model is worth its price depends in turn on which other models are
queried and how their outputs are combined, so the two decisions have to be
learned together from listings with known prices while keeping the average cost
within the budget. Problems of this kind have emerged with foundation models and
arise wherever such models are used to build predictors, from pricing products
\citep{vedula2025quantile,wang2026llp} to predicting clinical outcomes
\citep{guo2024multicenter}. We formulate them as a statistical learning problem
and develop methods and theory for it.

Several recent works use the task's own data to improve on the predictions of
pretrained models. \citet{li2026personalizing} and \citet{zhou2026blackbox}
correct the predictions of a single model with these data, and
\citet{kwon2026knowledge} correct several models separately and average them with
a convex weight. Our problem differs from theirs in three respects. First, they treat the
predictions as free and therefore never need to decide which models to query.
Second, they either use a single model or treat several models separately before
averaging, so they do not fit the outputs jointly with the features of the task
and cannot fully exploit how the models complement one another. Third, the most useful models may differ across inputs, and in our problem the queried models may vary accordingly. For instance, the numerical specifications may suffice to price some listings,
whereas a listing whose price depends on the processor or brand named in its
title may also need a language model.

We propose a framework that takes all three into account. It treats a procedure as a pair consisting of a query rule, which decides which models to query and may choose different models for different inputs, and a prediction rule, which combines the acquired outputs with the features of the task. Among the procedures whose average query cost does not
exceed a budget $B$, we seek the one with the smallest prediction risk.

We develop two methods, ACES-T and ACES-I, named for Affordable Combination of
External Signals at the Task level and at the Input level. Both learn from the
task's own data how much weight to give each acquired output, jointly with the
features, so that a scale bias in an output and the redundancy or
complementarity among outputs are taken into account. They differ in the query
rule. ACES-T queries the same collection of models for every input, which keeps
the cost of every input within the budget. Because the number of affordable collections, that is, sets of models whose total cost is within the budget, grows rapidly with the number of models, ACES-T minimizes the empirical risk over them by a
branch-and-bound search that avoids fitting every collection. ACES-I instead lets
the queried models vary across inputs, so that it can spend more on the inputs
that need more models. The value of querying a model then depends on which
models can still be queried afterwards, and ACES-I therefore learns when to stop
and which model to query next by backward Q-learning.

Our theory explains when several outputs help, what the budget costs, and
what is gained by letting the queried models vary across inputs. First, the risk of ACES-T depends on the approximation error of the affordable collection as a whole, so several outputs that are inaccurate on their own can jointly outperform the best single model. Second, under a H\"older model on a fixed input representation, ACES-T attains the minimax rate when the outcome depends on the outputs only through an affordable collection, in which case the budget does not slow the rate. Otherwise, the worst-case regret of every procedure within the budget relative to the best affordable collection decays only at a square-root rate, which ACES-T attains. Third, when different inputs prefer different collections, letting the queried models vary across inputs strictly improves on every affordable fixed collection, and ACES-I achieves this improvement once the sample is large enough.

In simulations and in a laptop price application with five foundation models queried in six configurations, both methods outperform existing methods that combine the predictions of
pretrained models with the task's own data. When one collection suits all inputs,
ACES-T nearly matches the accuracy of querying every model at a much lower cost
and ACES-I adds little. ACES-I improves on ACES-T when the most useful models vary across inputs.

\subsection{Related work}
\noindent\textbf{Routing and cascading.} Routing and cascading for large language
models also trade off response quality against query cost. A router selects one
model for each input before any response is generated
\citep{ong2025routellm,somerstep2025carrot}, whereas a cascade
queries models of increasing size and stops once the current response is judged
reliable \citep{chen2024frugalgpt,jitkrittum2023cascade}. In
both cases the final prediction is the response of one of the models. The outputs of the foundation models in our setting instead enter a prediction rule that is
fitted to the task's own data together with the features.

\noindent\textbf{Aggregation and ensembles.} Statistical aggregation combines
given predictors so as to perform nearly as well as the best one or the best
combination of them, with optimal rates
\citep{tsybakov2003optimal,lecue2014optimal}. These results
presume that every candidate is available at no cost, and the rate-optimal
procedures use all of them. Ensembles of large language models make the same
assumption \citep{jiang2023llmblender}. Among methods that account for cost,
the closest to ours is FrugalFoE \citep{wang2024fusing}. It queries experts
trained on different domains one at a time and fuses their outputs, aiming to match the accuracy of fusing all experts at a lower cost, and its fuser uses the input only through the expert outputs. Other
cost-aware ensembles select language models under a budget and combine the
selected outputs by a fixed rule such as voting
\citep{dai2024c2mabv,huang2025thriftllm}. How the prediction risk depends on the
sample size and on the budget when the combination itself is learned from the
task's own data remains open.

\noindent\textbf{Cost-sensitive feature acquisition.} The sequential queries of
ACES-I resemble cost-sensitive feature acquisition and prediction under
test-time budgets, in which a policy decides for each subject which feature to
measure next and when to stop
\citep{trapeznikov2013supervised,covert2023learning,zhou2026costoptimal}.
In this line of work, however, the acquired quantities are measurements of the
subject. Our problem instead involves predictions of the outcome produced by models
trained on other data, which may be biased for the task or redundant with one
another, so one must decide which predictions to buy and how to correct and
combine them with the features.

\section{Problem formulation}\label{sec:problem}

Let \(n\geq1\) and \(K\geq1\) denote the numbers of observations
from the task and of external foundation models, respectively, and write
\([k]=\{1,\ldots,k\}\) for any positive integer \(k\). We observe
\(O_i=(\bx_i,y_i)\), \(i\in[n]\), drawn independently from an unknown
distribution \(P\) of \((\bX,Y)\), where \(\bx_i\in\mathcal X\) is the input,
possibly with several modalities, and \(y_i\in\R\) is the outcome. Querying
model \(k\in[K]\) at \(\bx\) returns the prediction \(F_k(\bx)\in\R\), which may depend on only part of
the input (Remark~\ref{rem:external-library}). We write
\(\bF(\bx)=(F_1(\bx),\ldots,F_K(\bx))^\top\), \(F_{ik}=F_k(\bx_i)\), and
\(\bF_i=(F_{i1},\ldots,F_{iK})^\top\). A prespecified measurable feature map
\(\mathbf b_q:\mathcal X\to\R^q\), whose components include an intercept,
computes features from the input without querying any external model
(Remark~\ref{rem:local-representation}).

Querying model \(k\) costs a known amount \(c_k>0\). A collection is a set
\(A\subseteq[K]\) of models queried for the same input. We write \(\bF_A(\bx)\)
for the subvector of \(\bF(\bx)\) indexed by \(A\), with
\(\bF_{i,A}=\bF_A(\bx_i)\), and define the query cost of \(A\) as
\(C(A)=\sum_{k\in A}c_k\), so that \(C([K])\) is the cost of querying all
\(K\) models. The budget \(B\geq0\) bounds the average query cost per input at
deployment. Ties in every minimization below are broken by a prespecified
deterministic rule that favors the lower query cost.

A deployable procedure is a pair \(g=(\pi,m)\). The query rule \(\pi=\{d_A:A\subseteq[K]\}\) acquires outputs one at a time.
Starting from \(A=\varnothing\), it uses only the information acquired so far
to either stop or query an unqueried model, so that
\(d_A\{\mathbf b_q(\bx),\bF_A(\bx)\}\in\{\operatorname{stop}\}\cup([K]\setminus A)\)
and \(d_{[K]}=\operatorname{stop}\). Querying \(k\) enlarges the collection to \(A\cup\{k\}\), and
\(Q_\pi(\bx)\subseteq[K]\) denotes the collection at which \(\pi\) stops. The
prediction rule \(m=\{m_A:A\subseteq[K]\}\) forms the final prediction from
\(\mathbf b_q(\bx)\) and \(\bF_A(\bx)\) when \(\pi\) stops at \(A\). The
procedure returns
\begin{align*}
   g(\bx)=m_{Q_\pi(\bx)}\left\{\mathbf b_q(\bx), \bF_{Q_\pi(\bx)}(\bx)\right\}
\quad\text{at the cost}\quad C_g(\bx)=\sum_{k\in Q_\pi(\bx)}c_k. 
\end{align*}

Let \(\cG\) be a class of deployable procedures. For a prespecified loss \(\ell\), the population prediction risk of
\(g\in\cG\) is \(R_P^\ell(g)=\E_P\{\ell(Y,g(\bX))\}\) and its average query
cost is \(C_P(g)=\E_P\{C_g(\bX)\}\). The population target is
\begin{align}
\inf_{g\in\cG:\,C_P(g)\leq B}R_P^\ell(g).
\label{eq:population-target}
\end{align}
We assume that \(\cG\) contains the Zero-Query procedure, which stops at
\(A=\varnothing\) and queries no model. Because its cost is zero, the constraint in \eqref{eq:population-target} can always be met. Future observations follow the same distribution \(P\) as the training
data, so \(R_P^\ell\) and \(C_P\) describe accuracy and spending at
deployment. The budget constrains only the average cost over inputs, as when queries are billed by total usage, and the cost for an individual input may exceed \(B\).

For the \(n\) observations, we query all \(K\) models, which yields the training
sample \(\cD_n=\{(\bx_i,y_i,\bF_i):i\in[n]\}\). Because each row of \(\cD_n\) contains all
external outputs, any candidate procedure \(g=(\pi,m)\) can be replayed on the
training sample by revealing to \(\pi\) only the outputs that it queries. Its
empirical prediction loss and average query cost are
\begin{align}\label{eq:empirical-target}
\widehat R_n^\ell(g)=\frac{1}{n}\sum_{i=1}^n\ell\{y_i,g(\bx_i)\}, \qquad
\widehat C_n(g)=\frac{1}{n}\sum_{i=1}^n C_g(\bx_i) = \frac{1}{n}\sum_{i=1}^n
\sum_{k\in Q_\pi(\bx_i)}c_k.
\end{align}
Replacing the population risk and cost in \eqref{eq:population-target} by their
empirical counterparts in \eqref{eq:empirical-target} gives the sample analogue
\(\inf_{g\in\cG:\,\widehat C_n(g)\leq B}\widehat R_n^\ell(g)\).
Solving this problem over an unrestricted class \(\cG\) is impractical. The
number of query rules grows rapidly with \(K\), and a rule chosen to satisfy
the empirical constraint on the sample used to fit it need not satisfy the
budget at deployment. Section~\ref{sec:examples} therefore develops two
methods, ACES-T, which queries the same collection for every input, and
ACES-I, which queries a collection chosen for each input.

\begin{remark}
\label{rem:external-library}
Each index \(k\) refers to a model used with a fixed prompt, a fixed part of the
input and a fixed rule for converting its response into a number, so two indices
may refer to different models or to the same model with different prompts. A
vision-language model, for example, may receive only the image in the input, so
that \(F_k\) depends on \(\bx\) only through it. These choices are fixed before
the training outcomes are examined, so each \(F_k\) is a deterministic
measurable function of \(\bx\) whose relation to the outcome is a property of
\(P\) that can be learned from the training sample. Unlike in the single-model
formulations of \citet{li2026personalizing} and \citet{zhou2026blackbox}, the
input of each model need not coincide with \(\mathbf b_q(\bx)\).

\end{remark}

\begin{remark}
\label{rem:local-representation}
The feature map can be chosen to suit the data, and any map computed without
querying the models fits the framework. In the application of
Section~\ref{sec:real-data}, for example, each laptop is described by its numerical
specifications, a title and description, and a product image. Since text and
images cannot enter a regression directly, \(\mathbf b_q\) there consists of
spline bases of the numerical attributes, leading TF-IDF components of the text
and color moments of the image.
\end{remark}

\section{Method}\label{sec:examples}

\subsection{ACES-T}
\label{sec:task-level}

We first present ACES-T, which queries the same collection of models for
every input and chooses it among the affordable collections by the empirical
risk of the fitted prediction rules. Every input then incurs the cost of this
collection, which is within the budget.

\subsubsection{Prediction rules and the selection criterion}
\label{sec:task-method}

For each collection \(A\), ACES-T fits a prediction rule that combines the
features with the outputs in \(A\). The method and its search apply to any given function classes that are nested
in \(A\), and we use the linear classes
\begin{align*}
 \cH_A=\left\{\bx\mapsto \mathbf b_q(\bx)^\top\bgamma+\bF_A(\bx)^\top\btheta_A:
(\bgamma^\top,\btheta_A^\top)^\top\in\Theta_A\right\},
\end{align*}
where \(\bgamma\in\R^q\) and \(\btheta_A\in\R^{|A|}\) are the coefficient
vectors of the features and the external outputs, respectively. The domains
\(\Theta_A\subset\R^{q+|A|}\) are compact, convex and contain zero. They are
nested in the sense that padding \(\btheta_A\) with zeros maps \(\Theta_A\)
into \(\Theta_{A'}\) whenever \(A\subseteq A'\), so that \(\cH_A\subseteq\cH_{A'}\).

We use these classes because each output is itself a prediction of the
outcome. A coefficient on each output then corrects its scale bias, the
features absorb any part of its bias that they can represent, and fitting the
outputs of a collection jointly with the features accounts for their
redundancy and complementarity. The classes also cover nonlinear predictions,
since, with suitable features whose dimension grows with \(n\), the term
\(\mathbf b_q(\bx)^\top\bgamma\) approximates a H\"older function of a fixed input representation
(Theorem~\ref{thm:task-holder-rate}), and a basis expansion of the outputs,
such as splines of each output or products of outputs with features, makes the
prediction nonlinear in the outputs while keeping the class nested in \(A\).
Each fit involves only \(q+|A|\) coefficients, which keeps it stable, and the
risk bounds of Section~\ref{sec:theory} are stated in terms of this number,
with \(|A|\) replaced by the number of basis functions when the outputs are
expanded. Fully nonlinear classes, such as neural networks, can also be used,
and Section~\ref{sec:conclusion} discusses how they affect the search and the
risk bounds.

For a nonempty set \(I\subseteq[n]\) of training indices, the prediction rule for \(A\)
is fitted by minimizing the empirical risk over \(\Theta_A\),
\begin{align}
(\widehat\bgamma_{A,I},\widehat\btheta_{A,I})
\in\argmin_{(\bgamma^\top,\btheta_A^\top)^\top\in\Theta_A}
\frac1{|I|}\sum_{i\in I}
\ell\!\left\{y_i,\mathbf b_q(\bx_i)^\top\bgamma+\bF_{i,A}^\top\btheta_A\right\},
\label{eq:sample-terminal-fit}
\end{align}
which gives the fitted prediction rule
\(\widehat m_{A,I}\{\mathbf b_q(\bx),\bF_A(\bx)\}
=\mathbf b_q(\bx)^\top\widehat\bgamma_{A,I}+\bF_A(\bx)^\top\widehat\btheta_{A,I}\).
We write \(\widehat m_A=\widehat m_{A,[n]}\) for the fit on the full training
sample, \(\widehat R_n(A)\) for its attained minimum, and
\(m_A\in\argmin_{h\in\cH_A}\E_P\{\ell(Y,h(\bX))\}\) for the population prediction rule that \(\widehat m_A\) estimates, and set \(m=\{m_A:A\subseteq[K]\}\).

The affordable collections form \(\cA_B=\{A\subseteq[K]:C(A)\leq B\}\). Let
\(K_B=\max_{A\in\cA_B}|A|\) be the size of the largest affordable collection
and \(N_B=|\cA_B|\) the number of affordable collections. ACES-T selects
\begin{align}
\widehat A_B\in\argmin_{A\in\cA_B}\widehat R_n(A).
\label{eq:task-budget-estimator}
\end{align}
The resulting procedure \(\widehat g_{\rm task}\) queries \(\widehat A_B\) for
every input and predicts by
\(\widehat g_{\rm task}(\bx)
=\widehat m_{\widehat A_B}\{\mathbf b_q(\bx),\bF_{\widehat A_B}(\bx)\}\).
Restricting the minimization in \eqref{eq:task-budget-estimator} to
\(A=\varnothing\) or to \(|A|\leq1\) gives the Zero-Query and Single-Query
procedures \(\widehat g_{\rm zero}\) and \(\widehat g_{\rm single}\), the
benchmarks in Section~\ref{sec:theory-task}.

\subsubsection{Branch-and-bound search}
\label{sec:global-search}
Solving \eqref{eq:task-budget-estimator} directly requires fitting all
\(N_B\) affordable collections, and \(N_B\) grows quickly with \(K\), reaching
\(2^K\) when the budget covers every model. We instead solve it exactly by a
branch-and-bound search. It follows the
leaps-and-bounds idea of \citet{furnival1974regressions} for best-subset
regression, which uses the fit of a larger model to bound the fits of all its
submodels, and adds the budget as a constraint. The search keeps the best
collection found so far, called the incumbent, and partitions the remaining
candidates into nodes. A node is pruned if it contains no affordable
collection or if the fit of its largest collection, which bounds the
empirical risks of all its collections from below because the classes
\(\cH_A\) are nested, is worse than the incumbent. Supplementary~\ref{app:bb-example} describes the search in detail, shows
that it returns the minimizer in \eqref{eq:task-budget-estimator} whenever
each fit is computed exactly, and illustrates it with an example of \(K=4\)
models.

\subsection{ACES-I}
\label{sec:sample-level}

We next present ACES-I, which uses the same prediction classes but lets the
queried models vary across inputs. Given the outputs acquired so far, it
decides for each input whether to stop or to query another model, so that it
can spend more on the inputs that benefit from additional outputs while
keeping the average cost within the budget.

\subsubsection{Cost-penalized Bellman recursion}
\label{sec:sample-method}

After the models in \(A\subseteq[K]\) have been queried, the available information
is the state \(\bZ_A(\bx)=\{\mathbf b_q(\bx)^\top,\bF_A(\bx)^\top\}^\top\), and
we write \(\bZ_A=\bZ_A(\bX)\), so that a prediction rule for \(A\) is a
function of the state. If the queries stop at \(A\), ACES-I predicts by \(m_A\), the best rule in \(\cH_A\) over
all inputs. These rules are shared by all query
rules, and since every output is observed in the training sample, each of them
can be estimated once from all observations. With \(m\) fixed, ACES-I chooses the query rule to minimize the risk in
\eqref{eq:population-target} subject to the budget.

ACES-I replaces the budget constraint by the Lagrangian
\(R_P^\ell(g)+\eta C_P(g)\) with a multiplier \(\eta\geq0\). If the minimizer
has average cost \(B\) at some \(\eta_B\), it also solves the constrained
problem, and Section~\ref{sec:sample-validation-selection} chooses the
multiplier from data because \(\eta_B\) is unknown. For a given \(\eta\), the optimal query rule is characterized by a backward
recursion over the states. Let \(S_A^\eta=\ell\{Y,m_A(\bZ_A)\}+\eta C(A)\) be the stopping loss at \(A\), and let \(v_\eta^{\mathrm{stop}}(A,\bZ_A)=\E(S_A^\eta\mid\bZ_A)\) be the stopping value. The
value function \(v_A^\eta(\bZ_A)\) is the smallest conditional expected
stopping loss that can be attained from state \(A\). It equals the stopping
value at \(A=[K]\), and for \(A\subsetneq[K]\) it satisfies
\begin{align}
\label{eq:sample-exact-values}
v_A^\eta(\bZ_A)
&=\min\left\{v_\eta^{\mathrm{stop}}(A,\bZ_A),\min_{k\notin A}
\E[v_{A\cup\{k\}}^\eta(\bZ_{A\cup\{k\}})\mid\bZ_A]\right\},\notag\\
\Delta_{A,k}^{\eta,*}(\bZ_A)
&=\E[v_{A\cup\{k\}}^\eta(\bZ_{A\cup\{k\}})\mid\bZ_A]-v_\eta^{\mathrm{stop}}(A,\bZ_A),\quad k\notin A.
\end{align}
The continuation contrast \(\Delta_{A,k}^{\eta,*}\) is the difference between
the continuation value \(\E[v_{A\cup\{k\}}^\eta(\bZ_{A\cup\{k\}})\mid\bZ_A]\) and the stopping value at \(A\). The optimal query rule \(\pi_\eta\)
stops at \(A\) if no contrast is negative and otherwise queries a model with
the smallest contrast.

The contrasts involve the unknown value functions, but they can be written as
regressions of pseudo-outcomes on the states. Set
\(U_{[K]}^{\eta,*}=S_{[K]}^\eta\) and, for \(A\subsetneq[K]\) in decreasing
order of \(|A|\), define the pseudo-outcome
\(T_{A,k}^{\eta,*}=U_{A\cup\{k\}}^{\eta,*}-S_A^\eta\) and the realized value
\(U_A^{\eta,*}\) by
\begin{align}
\Delta_{A,k}^{\eta,*}(\bZ_A)=\E(T_{A,k}^{\eta,*}\mid\bZ_A),\quad
U_A^{\eta,*}=S_A^\eta+\min\{0,\min_{k\notin A}
\Delta_{A,k}^{\eta,*}(\bZ_A)\}.
\label{eq:sample-conditional-mean-recursion}
\end{align}
By the tower property and backward induction,
\(\E(U_A^{\eta,*}\mid\bZ_A)=v_A^\eta(\bZ_A)\), so the contrasts defined in
\eqref{eq:sample-conditional-mean-recursion} coincide with those in
\eqref{eq:sample-exact-values} as shown in
Supplementary~\ref{sec:pr-regression-verification}.


\subsubsection{Cross-fitted backward Q-learning}
\label{sec:sample-crossfit}

Following \eqref{eq:sample-conditional-mean-recursion}, ACES-I estimates the
contrasts in decreasing order of \(|A|\) by regressing the pseudo-outcomes on
the states, as in backward Q-learning for dynamic treatment regimes
\citep{murphy2005generalization}. However, the pseudo-outcomes are not
observed. ACES-I computes them from the fitted prediction rules and from the
contrasts fitted at larger states, and if these fits used the same
observations, the pseudo-outcomes would inherit their overfitting. To avoid this, ACES-I uses cross-fitting \citep{chernozhukov2018double}, with folds
\(I_1,\ldots,I_V\) that partition the training indices independently of the
data and complements \(I_{-v}=[n]\setminus I_v\). Since \(\eta_B\) is unknown,
the contrasts are fitted for every \(\eta\) in a prespecified finite grid \(\cE\subset[0,\infty)\). The following four steps
produce one candidate procedure \(\widehat g_\eta\) for each \(\eta\in\cE\).

\noindent\textbf{Step 1: Fit the prediction rules.}
For each collection \(A\), we fit \(\widehat m_{A,I_{-v}}\) by
\eqref{eq:sample-terminal-fit} for every fold \(v\), together with the
full-sample rule \(\widehat m_A\), which is shared across all \(\eta\in\cE\). If
\(B=0\), ACES-I returns the Zero-Query procedure based on
\(\widehat m_\varnothing\) and skips the remaining steps.

\noindent\textbf{Step 2: Fit temporary contrasts.}
For each \(\eta\in\cE\) and fold \(v\), we start from
\begin{align*}
    \widehat S_{i,A}^{\eta,(-v)}
=\ell\{y_i,\widehat m_{A,I_{-v}}(\bZ_A(\bx_i))\}+\eta C(A),
\qquad
\widehat U_{i,[K]}^{\eta,(-v)}=\widehat S_{i,[K]}^{\eta,(-v)},
\qquad i\in I_{-v}.
\end{align*}
We then proceed through \(|A|=K-1,\ldots,0\). For each \(k\notin A\), we
form the pseudo-outcomes
\(\widehat T_{i,A,k}^{\eta,(-v)}
=\widehat U_{i,A\cup\{k\}}^{\eta,(-v)}-\widehat S_{i,A}^{\eta,(-v)}\) and fit
the temporary contrast over a prespecified class \(\cQ_{A,k}\) by
\begin{align*}
\widehat\Delta_{A,k}^{\eta,(-v)}
\in\argmin_{f\in\cQ_{A,k}}
\frac1{|I_{-v}|}\sum_{i\in I_{-v}}
\ell_{\rm c}\bigl[\widehat T_{i,A,k}^{\eta,(-v)},
f\{\bZ_A(\bx_i)\}\bigr],
\end{align*}
where \(\ell_{\rm c}\) is a prespecified loss whose population minimizer is the conditional mean. After all contrasts at \(A\)
are fitted, we set
\(
\widehat U_{i,A}^{\eta,(-v)}
=\widehat S_{i,A}^{\eta,(-v)}+
\min\Bigl\{0,\min_{k\notin A}
\widehat\Delta_{A,k}^{\eta,(-v)}(\bZ_A(\bx_i))\Bigr\}
\).

\noindent\textbf{Step 3: Fit the final contrasts.}
For each held-out observation \(i\in I_v\), we keep the fits from Steps~1
and~2 fixed and evaluate the same recursion:
\[
\begin{alignedat}{2}
\widehat S_{i,A}^{\eta}
&=\ell\{y_i,\widehat m_{A,I_{-v}}(\bZ_A(\bx_i))\}+\eta C(A),
&\qquad
\widehat U_{i,[K]}^{\eta}&=\widehat S_{i,[K]}^{\eta},\\
\widehat U_{i,A}^{\eta}
&=\widehat S_{i,A}^{\eta}
+\min\Bigl\{0,\min_{k\notin A}
\widehat\Delta_{A,k}^{\eta,(-v)}(\bZ_A(\bx_i))\Bigr\},
&\qquad
\widehat T_{i,A,k}^{\eta}&=\widehat U_{i,A\cup\{k\}}^{\eta}-\widehat S_{i,A}^{\eta},
\quad k\notin A.
\end{alignedat}
\]
Each pseudo-outcome of observation \(i\) thus uses only fits that exclude
its fold. Pooling these pseudo-outcomes over the folds, we fit the final
contrasts
\begin{align*}
\widehat\Delta_{A,k}^{\eta}
\in\argmin_{f\in\cQ_{A,k}}
\frac1n\sum_{i=1}^n
\ell_{\rm c}\bigl[\widehat T_{i,A,k}^{\eta},
f\{\bZ_A(\bx_i)\}\bigr],
\qquad A\subsetneq[K],\ k\notin A.
\end{align*}

\noindent\textbf{Step 4: Form the query rule.} At each \(A\subsetneq[K]\), the fitted query rule \(\widehat\pi_\eta\) stops if
\(\min_{k\notin A}\widehat\Delta_{A,k}^\eta\{\bZ_A(\bx)\}\geq0\) and otherwise
queries a model \(k\notin A\) that minimizes
\(\widehat\Delta_{A,k}^\eta\{\bZ_A(\bx)\}\). Pairing it with
\(\widehat m=\{\widehat m_A:A\subseteq[K]\}\) gives the candidate procedure
\(\widehat g_\eta=(\widehat\pi_\eta,\widehat m)\).

\subsubsection{Selection by independent validation}
\label{sec:sample-validation-selection}

Section~\ref{sec:sample-crossfit} gives a candidate \(\widehat g_\eta\) for
each \(\eta\in\cE\), and repeating Steps~2--4 for several contrast learners
gives one candidate for each learner and each \(\eta\). These candidates and
the Zero-Query procedure are indexed by a finite set \(\cJ\) and written
\(\widehat g_\nu\), \(\nu\in\cJ\). ACES-I estimates their risks and costs on an independent validation
sample of \(n_{\rm val}\) observations drawn from \(P\), in which every output is observed. This sample can be obtained by randomly
splitting the available observations, as in Sections~\ref{sec:simulation}
and~\ref{sec:real-data}. The estimated risk and cost are
\(\widehat R_\nu^{\rm val}=\tfrac{1}{n_{\rm val}}\sum_{i=1}^{n_{\rm val}}
\ell\{y_i^{\rm val},\widehat g_\nu(\bx_i^{\rm val})\}\)
and
\(\widehat C_\nu^{\rm val}=\tfrac{1}{n_{\rm val}}\sum_{i=1}^{n_{\rm val}}
C_{\widehat g_\nu}(\bx_i^{\rm val})\),
respectively. It then returns \(\widehat g_{\rm input}=\widehat g_{\widehat\nu_B}\), where
\(\widehat\nu_B\) minimizes \(\widehat R_\nu^{\rm val}\) over \(\nu\in\cJ\) with
\(\widehat C_\nu^{\rm val}\leq B\).

\section{Theoretical analysis}
\label{sec:theory}

We write \(R(h)=\E_P\{\ell(Y,h(\bX))\}\) for the risk of a function \(h\) of the input, so that a procedure that queries a fixed collection and
predicts with \(h\) has risk \(R(h)\). For \(\bx\in\mathcal X\) and a prediction
value \(t\in\R\), \(R_{\bx}(t)=\E\{\ell(Y,t)\mid\bX=\bx\}\) is the conditional risk
of predicting \(t\) at \(\bx\). For a
function \(f\) of the input and \(1\leq r\leq\infty\), \(\|f\|_r\) denotes the
\(L_r(P_{\bX})\) norm, and \(L_2(P_{\bX})\) is the space of such functions with
\(\|f\|_2<\infty\).

\subsection{Reference class and loss conditions}
\label{sec:theory-targets}

Every procedure in Section~\ref{sec:examples} predicts with a function in some
class \(\cH_A\), so procedures that use different collections can be compared
through the smallest risk over a reference class that contains every
\(\cH_A\).

\begin{definition}[Reference class]
\label{def:black-box-assisted-class}
Let \(\cR\subseteq L_2(P_{\bX})\) be a closed subspace that contains each
component of \(\mathbf b_q\), and let \(F_k\in L_2(P_{\bX})\)
for each \(k\in[K]\). The reference class is
\(\cF_{\rm lin}=\{r+\boldsymbol\beta^\top\bF:
r\in\cR,\ \boldsymbol\beta\in\R^K\}\).
Assume that \(R\) attains its minimum over \(\cF_{\rm lin}\), fix a minimizer
\(h_0\) together with a representation
\(h_0=r_0+\boldsymbol\beta_0^\top\bF\), and write \(R_0=R(h_0)\).
\end{definition}

The space \(\cR\) specifies which functions of the input the reference class
may use besides the outputs. At one extreme, \(\cR\) is the span of the
features, so \(\cF_{\rm lin}\) is the linear span of \(\mathbf b_q\) and \(\bF\),
and \(h_0=\mathbf b_q^\top\bgamma_0+\bF^\top\boldsymbol\beta_0\) is the best
predictor linear in \(\mathbf b_q\) and \(\bF\). At the other extreme,
\(\cR=L_2(P_{\bX})\), so \(\cF_{\rm lin}=L_2(P_{\bX})\) and \(h_0\) minimizes
\(R\) over all functions of the input. Under squared loss, \(h_0\) is the
\(L_2\) projection of \(Y\) onto the linear span of \(\mathbf b_q(\bX)\) and
\(\bF(\bX)\) in the first case and equals \(\E(Y\mid\bX)\) in the second.
Because \(\cR\) contains the features, \(\cH_A\subseteq\cF_{\rm lin}\) for every
\(A\), so a procedure that queries a fixed collection has risk at least \(R_0\).
To compare fitted procedures with this lower bound, we impose two conditions.

\begin{assumption}
\label{ass:working-envelope}
(a) For some \(\bar t>0\), every function in every \(\cH_A\) takes values in
\(\mathcal T=[-\bar t,\bar t]\), \(\ell(y,\cdot)\) is continuous on \(\mathcal T\)
for every \(y\), and \(\E_P\{\ell(Y,0)\}<\infty\). (b) Almost surely, \(R_{\bX}\) is \(L_\ell\)-Lipschitz on \(\mathcal T\) for
some \(L_\ell\geq0\). (c) For some \(\sigma_\ell\geq0\) and all \(t,t'\in\mathcal T\),
\(\ell(Y,t)-\ell(Y,t')\) is conditionally sub-Gaussian given \(\bX\) with
variance proxy at most \(\sigma_\ell^2(t-t')^2\).
\end{assumption}

\begin{assumption}
\label{ass:conditional-curvature}
Almost surely \(h_0(\bX)\in\mathcal T\), and for some constants
\(0<c_\ell\leq C_\ell\),
\(c_\ell\|h-h_0\|_2^2\leq R(h)-R_0\leq C_\ell\|h-h_0\|_2^2\) for every
\(h\in\bigcup_{A\in\cA_B}\cH_A\).
\end{assumption}

Assumption~\ref{ass:working-envelope} bounds the predictions and the
fluctuations of loss differences, so that empirical risks can be compared
uniformly over the prediction classes. Assumption~\ref{ass:conditional-curvature}
states that the excess risk of a function in a prediction class is comparable to its squared distance from \(h_0\). Comparing collections then reduces to comparing the distances from \(h_0\) to their prediction classes.
Similar conditions appear in
\citet{vandegeer2008high,farrell2021deep},
and Supplementary~\ref{app:loss-verification} verifies both assumptions for
squared and logistic losses.

\subsection{Guarantees for ACES-T}
\label{sec:theory-task}

\subsubsection{Comparison with Zero-Query and Single-Query}
\label{sec:task-benchmarks}

To see when several outputs help, we compare ACES-T with the Zero-Query and
Single-Query procedures of Section~\ref{sec:task-method}, which use no output
and at most one output, respectively. Because they differ
from ACES-T only in the collections they may query, the comparison isolates the
value of querying a model and the value of combining several outputs. Their prediction classes and that of ACES-T are nested:
\(\cH_{\rm zero}=\cH_\varnothing\), \(\cH_{\rm single}\) is the union of
\(\cH_A\) over \(A\in\cA_B\) with \(|A|\leq1\), and \(\cH_{\rm task}\) is the
union of \(\cH_A\) over all \(A\in\cA_B\), so that
\(\cH_{\rm zero}\subseteq\cH_{\rm single}\subseteq\cH_{\rm task}\). For \(j\in\{{\rm zero},{\rm single},{\rm task}\}\), let
\(e_j=\inf_{h\in\cH_j}\|h_0-h\|_2\) be the distance from \(h_0\) to
\(\cH_j\). The nesting gives
\(e_{\rm task}\leq e_{\rm single}\leq e_{\rm zero}\).
Write \(V_{\rm task}=\inf_{h\in\cH_{\rm task}}R(h)\)
for the smallest risk over the affordable prediction class.

For \(\delta\in(0,1)\), the error of estimating the coefficients and
selecting a collection is controlled by \(\epsilon_{{\rm fit},n}\), given explicitly in
Supplementary~\eqref{eq:task-fast-radius}, with
\(\epsilon_{{\rm fit},n}\leq
Cn^{-1}\{(q+K_B)\log n+\log(N_B/\delta)\}\)
for \(n\geq2\), where \(C\) depends only on the constants in
Assumptions~\ref{ass:working-envelope}--\ref{ass:conditional-curvature}.

\Needspace{4\baselineskip}
\begin{theorem}[Prediction risk of ACES-T]
\label{thm:target-task-risk}
Under Assumptions~\ref{ass:working-envelope}--\ref{ass:conditional-curvature},
with probability at least $1-\delta$, simultaneously for
$j\in\{\mathrm{zero},\mathrm{single},\mathrm{task}\}$,
\begin{align}
c_\ell e_j^2\le R_P^\ell(\widehat g_j)-R_0 \le3C_\ell
e_j^2+\epsilon_{{\rm fit},n}.
\label{eq:target-task-fitted-bounds}
\end{align}
On the same event, for $j\in\{\mathrm{zero},\mathrm{single}\}$,
\begin{align}
R_P^\ell(\widehat g_j)-R_P^\ell(\widehat g_{\rm task}) \ge c_\ell e_j^2-3C_\ell
e_{\rm task}^2-\epsilon_{{\rm fit},n}.
\label{eq:target-task-dominance}
\end{align}
In particular, if
\(c_\ell e_{\rm single}^2>3C_\ell e_{\rm task}^2+\epsilon_{{\rm fit},n}\), then ACES-T has strictly lower risk than Zero-Query and Single-Query on this event.
\end{theorem}

Theorem~\ref{thm:target-task-risk} shows that the excess risk of each fitted
procedure is of the order of its squared approximation error \(e_j^2\), up to
the fitting error \(\epsilon_{{\rm fit},n}\). By \eqref{eq:target-task-dominance}, ACES-T improves on both
benchmarks when the best affordable collection approximates \(h_0\) much better
than the features combined with at most one output. This happens when the
outputs complement each other, as the following setting shows. Suppose that each
output has the form \(F_k=h_0+\xi_k\), where the bias \(\xi_k\) is orthogonal in
\(L_2(P_{\bX})\) to \(h_0\) and to each component of \(\mathbf b_q\). Suppose also
that \(h_0=\bF_A^\top\btheta_A\) for some \(A\in\cA_B\) and some \(\btheta_A\)
with \((\bzero_q^\top,\btheta_A^\top)^\top\in\Theta_A\), so that the biases of
the outputs in \(A\) cancel, and that \(\Theta_\varnothing\) contains the
coefficients of the \(L_2(P_{\bX})\) projection of \(h_0\) onto the span of the
components of \(\mathbf b_q\).

\begin{corollary}[Complementary outputs]
\label{cor:complementary-outputs}
In the setting above, \(e_{\rm task}=0\) and
\(
e_{\rm single}^2\geq\min_{k\in[K]}\frac{e_{\rm zero}^2\|\xi_k\|_2^2}{e_{\rm zero}^2+\|\xi_k\|_2^2}
\). Hence, under Assumptions~\ref{ass:working-envelope}--\ref{ass:conditional-curvature},
with probability at least \(1-\delta\),
\begin{align*}
R_P^\ell(\widehat g_{\rm single})-R_P^\ell(\widehat g_{\rm task})\geq c_\ell\min_{k\in[K]}\frac{e_{\rm zero}^2\|\xi_k\|_2^2}{e_{\rm zero}^2+\|\xi_k\|_2^2}-\epsilon_{{\rm fit},n}.
\end{align*}
\end{corollary}

If \(\|\xi_k\|_2\geq e_{\rm zero}\) for every \(k\), the lower bound on
\(e_{\rm single}^2\) is at least \(e_{\rm zero}^2/2\). A single output, even with
its best coefficient, then removes at most half of \(e_{\rm zero}^2\), whereas
the outputs in \(A\) together remove all of it. Scenario~1 of
Section~\ref{sec:simulation} uses this mechanism. The improvement concerns
prediction risk and requires neither coefficient nor support recovery.

\subsubsection{H\"older rate and minimax optimality}
\label{sec:task-holder}

The upper bound in \eqref{eq:target-task-fitted-bounds} has an approximation
term \(e_{\rm task}^2\) and a fitting term \(\epsilon_{{\rm fit},n}\) that grows
with \(q\). To obtain a rate, we bound \(e_{\rm task}^2\) under a H\"older
regression model based on a fixed representation of the input and choose \(q\)
to balance the two terms. For an integer \(p\geq1\), let \(\Psi:\mathcal X\to[0,1]^p\) be a measurable representation
that can be computed without querying any external model. Let
\(\Sigma(s,L_r)\) be the H\"older ball on \([0,1]^p\) with smoothness
\(s>0\) and radius \(L_r>0\), defined in
Supplementary~\eqref{eq:lb-holder}, and for a fixed \(\bar\beta>0\) let
\begin{align*}
\mathcal M_B=\left\{\widetilde r\circ\Psi+\boldsymbol\beta^\top\bF:
\widetilde r\in\Sigma(s,L_r),\quad
\operatorname{supp}(\boldsymbol\beta)\in\cA_B,\quad
\|\boldsymbol\beta\|_\infty\leq\bar\beta\right\}.
\end{align*}
We assume that \(\widetilde r\circ\Psi\in\cR\) for every
\(\widetilde r\in\Sigma(s,L_r)\), so that \(\mathcal M_B\subseteq\cF_{\rm lin}\).
Under squared loss, the outcome satisfies \(Y=h_0(\bX)+\varepsilon\), where
\(h_0=\widetilde r_0\circ\Psi+\boldsymbol\beta_0^\top\bF\in\mathcal M_B\) and
\(\varepsilon\) is conditionally centered and sub-Gaussian given \(\bX\) with
variance proxy \(0<\sigma^2<\infty\).

The rate uses the conditions in
Assumption~\ref{ass:feature-map} in the Supplementary Material:
the features approximate \(r_0=\widetilde r_0\circ\Psi\) at the
H\"older rate, the supremum and \(L_2(P_{\bX})\) norms are comparable
on the spans generated by the features and the outputs from any two
affordable collections, and the resulting approximation of \(h_0\) lies in an
affordable prediction class. For example, tensor-product B-splines on quasi-uniform
meshes of \([0,1]^p\), evaluated at \(\Psi(\bX)\), provide the required
approximation rate \citep{schumaker2007spline,huang2003local}.
The norm comparison is a standard condition for series estimation
\citep{newey1997convergence,belloni2015some}. We take
\(q=\lceil(nL_r^2/\sigma^2)^{p/(2s+p)}\rceil\) and assume that \(n\geq2\) and
\((q+K_B)\log(nN_B)\leq c_0n\) for a sufficiently small constant \(c_0>0\)
depending only on \(\kappa_Z\) and \(\nu_{\rm env}\) in
Assumption~\ref{ass:feature-map}.

For the minimax comparison, we restrict to the Gaussian special case in which \(\Psi(\bX)\sim{\rm Unif}([0,1]^p)\), \(\varepsilon\mid\bX\sim N(0,\sigma^2)\), and \(h_0\) ranges over
\(\mathcal M_B\). In this special case, the input distribution
\(P_{\bX}\), the representation \(\Psi\), the feature map \(\mathbf b_q\),
and the output functions \(\bF\) are fixed as \(h_0\) varies.
We assume that \(\E\{\bF(\bX)\bF(\bX)^\top\}=\mathbf I_K\),
that each \(\Theta_A\) contains
\(\{(\bzero_q^\top,\btheta_A^\top)^\top:\|\btheta_A\|_\infty\leq\bar\beta\}\), and
that Assumption~\ref{ass:feature-map} holds for every \(h_0\in\mathcal M_B\)
with the same constants. Let \(L_{\rm cost}\) be the number of distinct query costs, and let
\(\psi_n^\star=L_r^{2p/(2s+p)}\sigma^{4s/(2s+p)}n^{-2s/(2s+p)}
+\sigma^2(K_B+\log N_B)/n\).

\begin{theorem}[Rate and minimax optimality]
\label{thm:task-holder-rate}
Consider the H\"older regression model above.
\begin{enumerate}\renewcommand{\labelenumi}{(\roman{enumi})}\renewcommand{\theenumi}{(\roman{enumi})}
\item\label{item:holder-upper} Under Assumption~\ref{ass:feature-map}, with
\(C_{\rm app}\) and \(C_{\rm env}\) as defined there,
\begin{align}
\E\{R_P^\ell(\widehat g_{\rm task})-R_0\}\leq
C\bigl(\psi_n^\star+\sigma^2/n\bigr)+C_{\rm env}/n^2
\label{eq:task-holder-optimized-rate}
\end{align}
for a constant \(C=C(s,p,C_{\rm app})>0\).

\item\label{item:minimax-lower} In the Gaussian special case above, there exist
constants \(c=c(s,p)>0\), \(C=C(s,p,C_{\rm app})>0\) and \(n_0>0\) such that, for all \(n\geq n_0\),
\begin{align*}
\frac{c}{L_{\rm cost}}\psi_n^\star \leq\inf_{\widehat g}\sup_{h_0\in\mathcal M_B}
\E_{h_0}\{R_P^\ell(\widehat g)-R_0\}\leq\sup_{h_0\in\mathcal M_B}
\E_{h_0}\{R_P^\ell(\widehat g_{\rm task})-R_0\} \leq C\psi_n^\star,
\end{align*}
where \(\E_{h_0}\) is the expectation over the \(n\) observations under
\(h_0\), the threshold \(n_0\) depends on
\(s,p,L_r,\sigma,\bar\beta,K,C_{\rm env}\), and the infimum is over all procedures trained on these observations, including those that query all \(K\) models at deployment.
\end{enumerate}
\end{theorem}


In part~\ref{item:holder-upper}, balancing \(L_r^2q^{-2s/p}\) against
\(\sigma^2q/n\) gives the first term of \(\psi_n^\star\), the classical rate for
estimating a H\"older function of smoothness \(s\) on the
\(p\)-dimensional representation space
\citep{stone1982optimal}. The second term of \(\psi_n^\star\) is the cost of
estimating \(K_B\) output coefficients and selecting among \(N_B\) affordable
collections. Part~\ref{item:minimax-lower} shows that ACES-T attains the
minimax rate \(\psi_n^\star\) over \(\mathcal M_B\) when \(L_{\rm cost}\) and the
constants in Assumption~\ref{ass:feature-map} are bounded, even against
procedures that query all \(K\) models. For a single pretrained model, \citet{li2026personalizing} obtain a minimax rate whose first term has the same form, with the smoothness of the bias of its predictions in place of that of
\(\widetilde r_0\), and whose second term \(\sigma^2/n\) is the cost of estimating an
offset. With one affordable output, \(K_B+\log N_B\) is bounded, and our second
term is also of order \(\sigma^2/n\).


\subsubsection{Sharp regret and the price of a binding budget}
\label{sec:task-sharp-regret}

We now allow \(h_0\) to depend on more outputs than any
affordable collection contains. Then \(h_0\) may lie outside every affordable
prediction class, so \(V_{\rm task}\) can exceed \(R_0\) by an approximation error incurred by every predictor in \(\cH_{\rm task}\). Theorem~\ref{thm:target-task-risk} does not show that the risk of ACES-T
approaches \(V_{\rm task}\), because
\(V_{\rm task}-R_0\leq C_\ell e_{\rm task}^2\) by
Assumption~\ref{ass:conditional-curvature} whereas its upper bound multiplies
\(e_{\rm task}^2\) by \(3C_\ell\). We therefore measure ACES-T by its regret
\(R_P^\ell(\widehat g_{\rm task})-V_{\rm task}\), which separates the error of
fitting within a collection from the error of selecting a collection. For \(A\in\cA_B\), let
\(V(A)=\inf_{h\in\cH_A}R(h)\), so that the optimal affordable collections
form the set \(\cA_B^*=\{A\in\cA_B:V(A)=V_{\rm task}\}\), which may contain
more than one element. Throughout this subsection, we use squared loss and the regression model
\(Y=h_0(\bX)+\varepsilon\), where \(\varepsilon\) is conditionally centered
and sub-Gaussian with variance proxy \(\sigma^2\). The population prediction
rule \(m_A\) of Section~\ref{sec:task-method} is then the
\(L_2(P_{\bX})\) projection of \(h_0\) onto \(\cH_A\). Fix \(A^\dagger\in\cA_B^*\) and write \(\upsilon_A=\|m_A-m_{A^\dagger}\|_2\). Let \(\kappa_4\) be a constant such that
\(\|f\|_4\leq\kappa_4\|f\|_2\) for every \(f\) in the linear
span of \((\mathbf b_q,\bF)\) and every residual \(h_0-m_A\) with \(A\in\cA_B\).
Define \(\rho_B=\max_{A\in\cA_B}\upsilon_A\) and
\(\sigma_{\rm sel}^2=\sigma^2+\kappa_4^4\max_{A\in\cA_B}\|h_0-m_A\|_2^2\), and let
\(\kappa_B\) be the maximum of \(\upsilon_A^2/\{V(A)-V_{\rm task}\}\) over
\(A\in\cA_B\setminus\cA_B^*\), or zero if every affordable collection is optimal.

For the lower bound, we consider an independent-sign setting in which every
output carries signal but an unknown group of \(K_B\) outputs has slightly
larger coefficients. The outputs \(F_1,\ldots,F_K\) are independent
Rademacher variables, every query costs one, \(B=K_B\) for an integer
\(1\leq K_B\leq K/32\), the feature map is \(b_1\equiv1\), and
\(\Theta_A=[-2\tau_{\rm sig},2\tau_{\rm sig}]^{1+|A|}\). Let
\(A_1,\ldots,A_{N_{\rm pack}}\) be the \(K_B\)-subsets of \([K]\) constructed in
Lemma~\ref{lem:PI-packing-testing}, any two of which differ in at least
\(K_B/2\) elements. Under hypothesis \(A\in\{A_1,\ldots,A_{N_{\rm pack}}\}\),
\(Y=\boldsymbol\beta^\top\bF+\varepsilon\), where
\(\boldsymbol\beta=\tau_{\rm sig}\bone_{[K]}+a_{\rm sig}\bone_A\)
and \(\varepsilon\sim N(0,\sigma^2)\) is independent of \(\bX\).
Here \(\bone_A\in\{0,1\}^K\) is the indicator vector of \(A\), \(0<\tau_{\rm sig}\leq\sigma/\sqrt{K+3K_B}\) and \(a_{\rm
sig}=\sigma\{\log N_{\rm pack}/(4nK_B)\}^{1/2}\leq\tau_{\rm sig}\). An oracle
spends its budget on the group \(A\), whereas a learned procedure must identify \(A\) from the training sample.
The smallest risk of any procedure \(g\) with \(C_P(g)\leq B\) equals
\(V_{\rm task}=\sigma^2+(K-K_B)\tau_{\rm sig}^2\).

\begin{theorem}[Sharp regret and the price of a binding budget]
\label{thm:task-sharp-regret}
Consider the regression model above under squared loss.
\begin{enumerate}\renewcommand{\labelenumi}{(\roman{enumi})}\renewcommand{\theenumi}{(\roman{enumi})}
\item\label{item:sharp-upper} Suppose that Assumption~\ref{ass:working-envelope}(a) holds,
\(\|h_0\|_\infty\leq\bar t\) and \(\sigma^2\leq\bar t^2\). Then for \(n\geq2\), with probability at least
\(1-\delta\),
\begin{align}
R_P^\ell(\widehat g_{\rm task})-V_{\rm task} \leq C\Bigl[
\min\bigl\{\rho_B\sigma_{\rm sel}\sqrt{\frac{\log(4N_B/\delta)}n},\
\kappa_B\sigma_{\rm sel}^2\frac{\log(4N_B/\delta)}n\bigr\}
+\epsilon_{{\rm fit},n}\Bigr],
\label{eq:task-sharp-regret}
\end{align}
where \(C>0\) is a universal constant.
\item\label{item:budget-lower} In the independent-sign setting, every procedure \(\widehat g\) trained on
\(\cD_n\) with \(C_P(\widehat g)\leq B\) almost surely satisfies
\begin{align}
\max_{A\in\{A_1,\ldots,A_{N_{\rm pack}}\}} \E_A\{R_P^\ell(\widehat g)-V_{\rm
task}\} \geq c\,\rho_B\sigma\sqrt{\log N_B/n}
\label{eq:task-budget-lower}
\end{align}
where \(\E_A\) is the expectation over \(\cD_n\) under hypothesis \(A\) and
\(c>0\) is a universal constant.
\end{enumerate}
\end{theorem}

The minimum in \eqref{eq:task-sharp-regret} is the error of selecting a
collection. If an affordable prediction
class contains \(h_0\), then \(\upsilon_A^2=V(A)-V_{\rm task}\), so \(\kappa_B=1\) whenever a
suboptimal collection exists, and the selection error is of order
\(\log N_B/n\). The same order holds whenever \(\kappa_B\) is bounded, even if
no affordable prediction class contains \(h_0\). The square-root branch matters only when
collections with nearly equal risks give substantially different predictions.

Part~\ref{item:budget-lower} shows that a binding budget can create exactly
this situation, and then no procedure within the budget avoids the square-root
rate. Unlike in Theorem~\ref{thm:task-holder-rate}, \(h_0\) lies outside every
affordable prediction class, because every output carries signal and an unknown
group \(A\) of \(K_B\) outputs carries slightly more. Querying a group disjoint from
\(A\) instead of \(A\) raises the risk by
\(K_B(2\tau_{\rm sig}a_{\rm sig}+a_{\rm sig}^2)\), which is linear in the extra
signal \(a_{\rm sig}\), while \(a_{\rm sig}\) is too small for the training
sample to identify \(A\) reliably. This loss is of the order of the right side
of \eqref{eq:task-budget-lower}. If \(\tau_{\rm sig}\) were zero, so that
\(h_0\in\cH_A\), the loss would be \(K_Ba_{\rm sig}^2\), of order
\(\sigma^2\log N_{\rm pack}/n\). Without a budget, model-selection aggregation
averages the predictions of competing collections and attains a
\(\log N_B/n\) remainder \citep{tsybakov2003optimal,lecue2014optimal}, but it
needs the outputs of all of them. Under a budget, a procedure must pick one.
The bound applies to every procedure within the budget, including those whose
queries depend on the input. Since \(\sigma\leq\sigma_{\rm sel}\leq2\sigma\) in
the independent-sign setting, ACES-T attains this rate up to
\(\epsilon_{{\rm fit},n}\).

\subsection{Guarantees for ACES-I}
\label{sec:structural-sample-risk}

\subsubsection{Population gain from input-dependent queries}
\label{sec:sample-observable-heterogeneity}

Holding the population
prediction rules \(m=\{m_A\}\) of Section~\ref{sec:task-method} fixed, define
\(V_{\rm input}(B)=\inf_{g=(\pi,m):\,C_P(g)\leq B}R_P^\ell(g)\)
and \(\Gamma_B=V_{\rm task}-V_{\rm input}(B)\),
where the infimum is over measurable deterministic query rules \(\pi\) whose
decision at each state depends only on \(\bZ_A\). Querying an affordable fixed
collection is such a rule, so \(\Gamma_B\geq0\), and \(\Gamma_B\) is the gain
from input-dependent queries. For \(\eta\geq0\), let \(\pi_\eta\) be the
population query rule of Section~\ref{sec:sample-method} with multiplier
\(\eta\), and let \(c(\eta)=\E C\{Q_{\pi_\eta}(\bX)\}\) be its average query
cost. By Bellman optimality, \((\pi_\eta,m)\) minimizes
\(R_P^\ell(g)+\eta C_P(g)\) over such rules. The cost curve \(c(\eta)\) is
nonincreasing in \(\eta\) but need not be continuous. To compare the fitted
procedure with a population procedure that spends exactly the budget, we impose the
following condition.

\begin{assumption}[Lipschitz cost curve]
\label{ass:input-dependent-collections}
There exist constants \(0<\varrho_\eta<\eta_-<\eta_+<\infty\) and
\(0\leq L_C<\infty\) such that
\(|c(\eta)-c(\eta')|\leq L_C|\eta-\eta'|\) for all
\(\eta,\eta'\in[\eta_--\varrho_\eta,\eta_++\varrho_\eta]\), and
\(c(\eta_-)\geq B>c(\eta_+)\).
\end{assumption}

Assumption~\ref{ass:input-dependent-collections} requires the average cost of
\(\pi_\eta\) to change gradually with \(\eta\) near
\(I_\eta=[\eta_-,\eta_+]\), and the budget to lie within the range of costs
attained there. It is natural when different inputs stop querying at different
multipliers, so that no positive fraction of them changes action at the same
\(\eta\). Under this assumption, there exists \(\eta_B\in I_\eta\) with
\(c(\eta_B)=B\), and the procedure \((\pi_{\eta_B},m)\) attains \(V_{\rm input}(B)\),
because it minimizes \(R_P^\ell(g)+\eta_BC_P(g)\) and spends exactly the budget.
The constant \(L_C\) also converts errors in the fitted contrasts and in the
multiplier grid into errors in average cost.
Supplementary~\ref{app:cost-transition-verification} gives sufficient
conditions in terms of the distribution of the multipliers at which collections
tie.

The gain \(\Gamma_B\) is positive when different inputs prefer different
collections. Let \(\bar v_\eta^{\mathrm{stop}}(A,\bz)=\E(S_A^\eta\mid\bZ_\varnothing=\bz)\) be
the conditional mean of the stopping loss at \(A\) given the initial state. The
next assumption requires different collections to be uniquely preferred at two
points of the support.

\begin{assumption}[Collection heterogeneity]
\label{ass:observable-heterogeneity}
There exist two points \(\bz_1,\bz_2\) in the support of \(P_{\bZ_\varnothing}\)
such that, for each \(A\), \(\bar v_{\eta_B}^{\mathrm{stop}}(A,\cdot)\) has a version that is
continuous in neighborhoods of both points, and \(\bar
v_{\eta_B}^{\mathrm{stop}}(\cdot,\bz_i)\) admits a unique minimizer \(A_i^\circ\) for \(i=1,2\), with
\(A_1^\circ\neq A_2^\circ\).
\end{assumption}

\begin{proposition}[Positive gain from heterogeneity]
\label{prop:population-routing-advantage}
Suppose that Assumptions~\ref{ass:working-envelope},
\ref{ass:input-dependent-collections} and \ref{ass:observable-heterogeneity}
hold. Then each \(\bz_i\) has a neighborhood \(\mathcal U_i\) with \(\mathbb P(\bZ_\varnothing\in\mathcal U_i)>0\) on which \(\bar v_{\eta_B}^{\mathrm{stop}}(A_i^\circ,\bz)\) is
smaller than \(\bar v_{\eta_B}^{\mathrm{stop}}(A,\bz)\) for every other collection \(A\) by at
least some \(\gamma_i>0\). Consequently,
\begin{align*}
\Gamma_B\geq\min_{i=1,2}\gamma_i\,\mathbb P(\bZ_\varnothing\in\mathcal U_i)>0, \qquad R(h)-V_{\rm
input}(B)\geq\Gamma_B \quad\text{for every }h\in\cH_{\rm task}.
\end{align*}
The gain also persists against input-independent randomization over fixed
collections. If a procedure queries \(A^{(l)}\subseteq[K]\) with probability
\(\lambda_l\) independently of the input and predicts with \(m_{A^{(l)}}\), then
its risk exceeds \(V_{\rm input}(B)\) by at least the same amount whenever its average cost
\(\sum_l\lambda_lC(A^{(l)})\) is at most \(B\). This holds even if some
\(A^{(l)}\) are unaffordable.
\end{proposition}

The lower bound on \(\Gamma_B\) in Proposition~\ref{prop:population-routing-advantage} grows with the probability of each region and the strength of the preference within it. The last statement
attributes the gain to adapting the collection to the input: randomizing over
fixed collections, even unaffordable ones, can spend the budget just as fully
and still falls short by the same amount.


\subsubsection{Prediction risk of the fitted procedure}
\label{sec:sample-fitted-transfer}

When several contrast learners are used, the validation step of
Section~\ref{sec:sample-validation-selection} chooses among the candidates of
all of them. We fix one prespecified learner as a reference, write \(\widehat\Delta_{A,k}^\eta\) for its fitted contrasts, and consider, for tolerances \(\epsilon_{m,n}\) and \(\epsilon_{\Delta,n}\), the event
\begin{align}
\max_{A\subseteq[K]}\|\widehat m_A-m_A\|_\infty \leq\epsilon_{m,n},\quad
\max_{\substack{\eta\in\cE\cap I_\eta\\A\subsetneq[K],\ k\notin A}}
\|\widehat\Delta_{A,k}^\eta-\Delta_{A,k}^{\eta,*}\|_\infty
\leq\epsilon_{\Delta,n}.
\label{eq:sample-exact-training-event}
\end{align}
The other learners only add candidates and need not
satisfy \eqref{eq:sample-exact-training-event}. On this event, for each \(\eta\in\cE\cap I_\eta\), the query rule
fitted by the reference learner, paired with the population prediction rules,
has a value of
\(R_P^\ell(g)+\eta C_P(g)\) that exceeds that of \((\pi_\eta,m)\) by at most
\(r_n=2K\epsilon_{\Delta,n}\), and under
Assumption~\ref{ass:input-dependent-collections} its average cost differs from
\(c(\eta)\) by at most $\tau_n=\sqrt{2L_Cr_n}+2r_n/\varrho_\eta$.

The candidates cover only the multipliers in the grid \(\cE\), so the
multiplier \(\eta_B\) is approximated up to the grid spacing. When \(\cE\cap I_\eta\ne\varnothing\), we define the mesh
\(\omega_\eta\) as the largest gap between consecutive points of
\((\cE\cap I_\eta)\cup\{\eta_-,\eta_+\}\).
For example, an equally spaced grid of \(\lceil\sqrt n\rceil+1\) points on
\(I_\eta\) gives \(\omega_\eta=O(n^{-1/2})\). Finally, the candidates \(\widehat g_\nu\),
\(\nu\in\cJ\), are fixed before validation. Define \(\epsilon_{C,n_{\rm val}}=C([K])\sqrt{\log\{8|\cJ|/\delta\}/(2n_{\rm val})}\),
where \(|\cJ|\) is the number of candidates, and let \(\mathcal E_{\rm input}\),
given explicitly in Supplementary~\eqref{eq:validation-radii}, collect the errors
of fitting, of the multiplier grid and of validation. It satisfies
\(
\mathcal E_{\rm input}\leq C\bigl[\epsilon_{m,n}+\epsilon_{\Delta,n}^{1/2}+\omega_\eta
+\{\log(|\cJ|/\delta)/n_{\rm val}\}^{1/2}\bigr]
\)
whenever \(\epsilon_{\Delta,n}\leq1\) and \(\log(|\cJ|/\delta)\leq n_{\rm val}\),
where \(C\) is a positive constant depending only on \(K\), \(C([K])\), \(\eta_+\), \(L_C\), \(\varrho_\eta\), and the
constants in Assumption~\ref{ass:working-envelope}.
\begin{assumption}[Fitting accuracy and budget margin]
\label{ass:fitting-margin}
The event \eqref{eq:sample-exact-training-event} has probability at least
\(1-\delta/2\), the grid satisfies \(\cE\cap I_\eta\ne\varnothing\), and the budget leaves a
margin for the errors of fitting and validation,
\(B-\tau_n-\epsilon_{C,n_{\rm val}}\geq c(\eta_+)+L_C\omega_\eta\).
\end{assumption}

\begin{theorem}[Prediction risk of ACES-I]
\label{thm:validation-selector}
Suppose that Assumption~\ref{ass:working-envelope} holds.
\begin{enumerate}\renewcommand{\labelenumi}{(\roman{enumi})}\renewcommand{\theenumi}{(\roman{enumi})}
\item\label{item:routing-cost-bound} With probability at least \(1-\delta/2\),
\(C_P(\widehat g_{\rm input})\leq B+\epsilon_{C,n_{\rm val}}\).
\item\label{item:routing-risk-bound} If Assumptions~\ref{ass:input-dependent-collections}
and~\ref{ass:fitting-margin} also hold, then with probability at least
\(1-\delta\), the bound in part~\ref{item:routing-cost-bound} holds and, for
every \(h\in\cH_{\rm task}\),
\begin{align}
R_P^\ell(\widehat g_{\rm input})
\leq V_{\rm input}(B)+\mathcal E_{\rm input},\quad R(h)-R_P^\ell(\widehat g_{\rm input})
\geq\Gamma_B-\mathcal E_{\rm input}.
\label{eq:sample-structural-advantage}
\end{align}
\end{enumerate}
\end{theorem}

Because \eqref{eq:sample-structural-advantage} holds for
every \(h\in\cH_{\rm task}\), it applies to the fitted Zero-Query,
Single-Query, and ACES-T procedures. Whenever \(\Gamma_B>\mathcal E_{\rm
input}\), ACES-I therefore improves on all three by at least
\(\Gamma_B-\mathcal E_{\rm input}\), while its average cost exceeds the budget
by at most \(\epsilon_{C,n_{\rm val}}\). Under the conditions of
Proposition~\ref{prop:population-routing-advantage}, \(\Gamma_B>0\), so this
holds once the errors collected in \(\mathcal E_{\rm input}\) are small enough. The bound on \(\mathcal E_{\rm input}\) shows that errors in the fitted
contrasts enter through their square root, because they shift the average cost
of the fitted query rule.
Supplementary~\ref{app:extensions-verifications} bounds \(\epsilon_{m,n}\) and
\(\epsilon_{\Delta,n}\) explicitly when the contrasts are fitted by least
squares.

\section{Simulation study}\label{sec:simulation}

We evaluate the finite-sample performance of ACES-T and ACES-I in two
scenarios. In Scenario~1 one collection of models suits all inputs, and in Scenario~2 the most useful models vary across inputs. Both scenarios draw the
features \(X_1,\ldots,X_p\) independently from the uniform distribution on
\((-1,1)\), with \(p=6\) in Scenario~1 and \(p=7\) in Scenario~2, and use
\(\mathbf b_q(\bX)=(1,\bX^\top)^\top\). Let
\(\varphi_d=\sqrt{2d+1}L_d\), where \(L_d\) is the degree-\(d\) Legendre
polynomial. The scenarios also share the linear component
\(\mathbf b_q(\bX)^\top\bgamma\), with
\(\bgamma=(1,0.5,-0.5,0.3,0,\ldots,0)^\top\in\R^q\), and the nonlinear
components
\(Z_r=3^{-1/2}\sum_{h=1}^3
\varphi_{r+2}(X_{2h-1})\varphi_{r+2}(X_{2h})\),
\(r=0,1\).
The noise \(\varepsilon\sim N(0,0.25)\) is independent
of \(\bX\), and we set \(B=0.125\).

\noindent\textbf{Scenario 1.} The outcome is
\(Y=\mathbf b_q(\bX)^\top\bgamma+Z_0+\varepsilon\), and the five external
outputs are
\begin{align*}
F_k(\bX)&=\E(Y\mid\bX)+a_{k1}\varphi_4(X_1)+a_{k2}\varphi_4(X_2)+0.25\varphi_5(X_k),
\qquad k=1,2,3,\\
F_4(\bX)&=\E(Y\mid\bX)+0.65\varphi_6(X_1),\quad
F_5(\bX)=\E(Y\mid\bX)+0.75\varphi_6(X_2),
\end{align*}
where \((a_{11},a_{12})=(2.8,0.6)\), \((a_{21},a_{22})=(-0.8,3.2)\) and
\((a_{31},a_{32})=(-2.2,-2.6)\), and the query costs are
\(\mathbf c=(0.02,0.025,0.03,0.10,0.12)^\top\). Each of \(F_1\), \(F_2\) and
\(F_3\) has a large bias, but their biases in \(\varphi_4(X_1)\) and
\(\varphi_4(X_2)\) cancel in a suitable linear combination.

\noindent\textbf{Scenario 2.} The feature \(X_7\) governs the relative
strength of the nonlinear components through \(w=\{1+\exp(-16X_7)\}^{-1}\),
and the outcome is \(Y=\mathbf b_q(\bX)^\top\bgamma+(1-w)Z_0+wZ_1+\varepsilon\).
For \(k=1,2,3\), the inexpensive outputs are
\begin{align*}
F_k(\bX)&=\E(Y\mid\bX)-0.85wZ_1+a_{k1}\varphi_4(X_1)
+a_{k2}\varphi_4(X_2)+0.25\varphi_5(X_k),\\
F_{k+3}(\bX)&=\E(Y\mid\bX)-0.85(1-w)Z_0+a_{k1}\varphi_4(X_3)
+a_{k2}\varphi_4(X_4)+0.25\varphi_5(X_{k+3}),
\end{align*}
and the remaining outputs are
\(F_7(\bX)=\E(Y\mid\bX)+0.65\varphi_6(X_1)\)
and \(F_8(\bX)=\E(Y\mid\bX)+0.75\varphi_6(X_2)\),
with query costs
\(\mathbf c=(0.02,0.025,0.03,0.02,0.025,0.03,0.10,0.12)^\top\). Thus
\(F_1\), \(F_2\) and \(F_3\) retain only \(15\%\) of \(wZ_1\), and \(F_4\),
\(F_5\) and \(F_6\) retain only \(15\%\) of \((1-w)Z_0\). Because \(Z_0\)
dominates \(\E(Y\mid\bX)\) when \(X_7<0\) and \(Z_1\) dominates when \(X_7>0\),
inputs with \(X_7\leq0\) prefer the collection \(\{1,2,3\}\) and inputs with
\(X_7>0\) prefer \(\{4,5,6\}\). Each of these collections costs \(0.075\), but
their union costs \(0.15>B\), so no affordable fixed collection contains both.

\begin{figure}[htbp]
	\centering
	\includegraphics[width=0.8\textwidth]{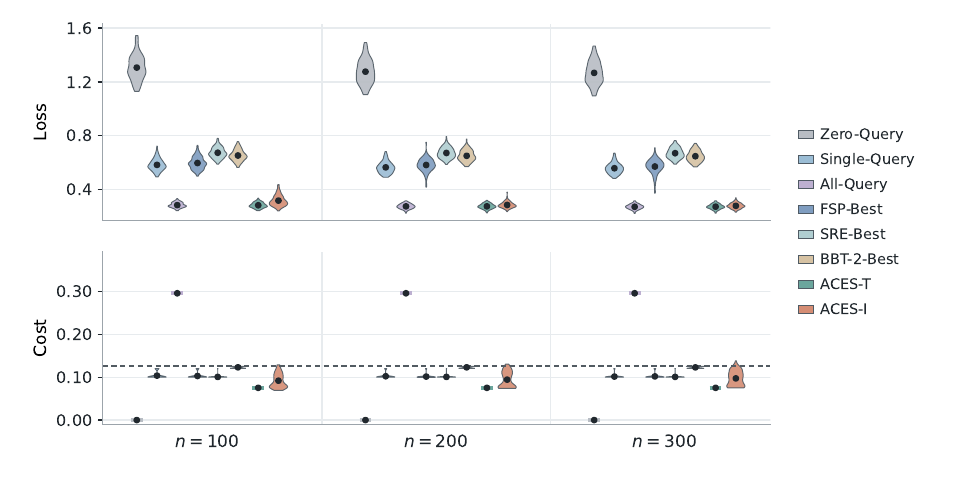}
	\caption{Test MSE (upper panel) and query cost (lower panel) in
		Scenario~1. Violins show distributions over 100 replications, dots
		mark means, horizontal bars mark quantities that are constant across
		replications, and the dashed line marks \(B\).}
	\label{fig:sim1-performance}
\end{figure}

For each scenario and \(n\in\{100,200,300\}\), we generate 100 replications,
each with training and validation samples of size \(n\) and a test sample of
size 500, so that all methods use the same \(2n\) observations. We
compare ACES-T and ACES-I with Zero-Query, Single-Query and All-Query, which, like
ACES-T, are fitted on the pooled training and validation samples. All-Query queries every model and serves as an over-budget reference. We also include three methods from the literature, which, like ACES-I, are fitted on the training sample and tuned on the validation sample: FSP applies the local correction of \citet{li2026personalizing} with a fixed training sample and
validation-tuned neighborhoods, SRE applies the residual estimation and safety
construction of \citet{zhou2026blackbox}, and BBT-2 adapts the combination of
\citet{kwon2026knowledge} to outputs observed on the same sample by
combining two raw or residual-corrected outputs with a validation-tuned convex
weight. FSP and SRE are fitted for each external model and BBT-2 for each
affordable pair, and FSP-Best, SRE-Best and BBT-2-Best report the fit with the
lowest test mean squared error (MSE), a choice that favors these benchmarks.
Both ACES methods fit their prediction rules by unpenalized least squares
on \(\mathbf b_q(\bx)\) and the acquired outputs. Supplementary
Proposition~\ref{prop:unconstrained-ols-bridge} gives conditions under which
this fit coincides with the fit over \(\Theta_A\) in
\eqref{eq:sample-terminal-fit}. ACES-I uses five-fold
cross-fitting and fits each contrast by least squares over constants and
single-split trees with two leaf means, where candidate splits use a feature at
\(\{-0.5,0,0.5\}\) or an acquired output at \(\{-4,-2,0,2,4\}\), and its
multiplier grid consists of zero and 81 logarithmically spaced values from
\(10^{-3}\) to \(10^3\). Supplementary Lemma~\ref{lem:finite-stump-errors} bounds the estimation
error of this contrast class.

\begin{figure}[htbp]
\centering
\includegraphics[width=0.85\textwidth]{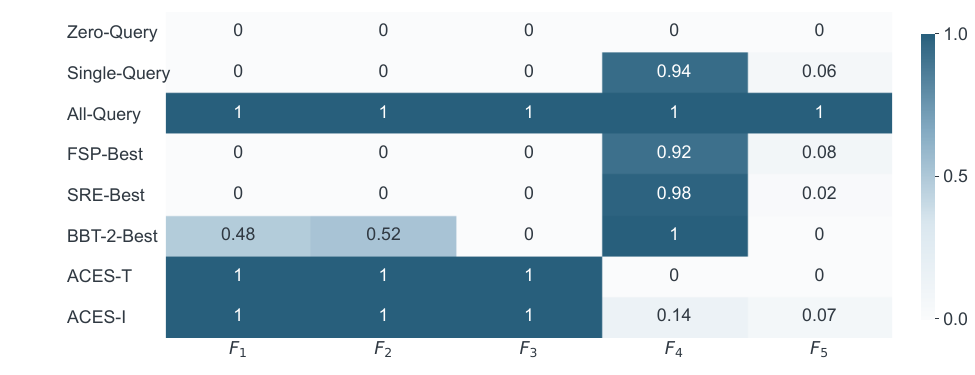}
\caption{Fraction of test inputs on which each model is queried in
Scenario~1 at \(n=300\), averaged over replications.}
\label{fig:sim1-calls}
\end{figure}

Figure~\ref{fig:sim1-performance} shows that both ACES methods have
lower mean test MSE than all benchmarks except All-Query at every sample size. At \(n=300\), ACES-T and ACES-I attain mean test MSEs of \(0.275\)
and \(0.281\) at mean query costs of \(0.075\) and
\(0.097\), whereas Single-Query and the three methods from the literature have mean test MSEs between \(0.560\) and \(0.672\) at costs between \(0.100\) and
\(0.123\). Figure~\ref{fig:sim1-calls} explains the difference. These benchmarks query model 4 on \(92\%\) to \(100\%\) of test inputs, while ACES-T queries models
1, 2 and 3 for every input. ACES-T thus nearly matches the \(0.274\) mean test MSE of All-Query at about one quarter of its \(0.295\) cost. ACES-I relies on the same
three models but also queries models 4 and 5 on about \(14\%\) and \(7\%\) of
test inputs. These extra queries raise its mean cost, and its mean test MSE stays above that of ACES-T at each sample size, although the gap narrows as \(n\) increases.

\begin{figure}[htbp]
	\centering
	\includegraphics[width=0.8\textwidth]{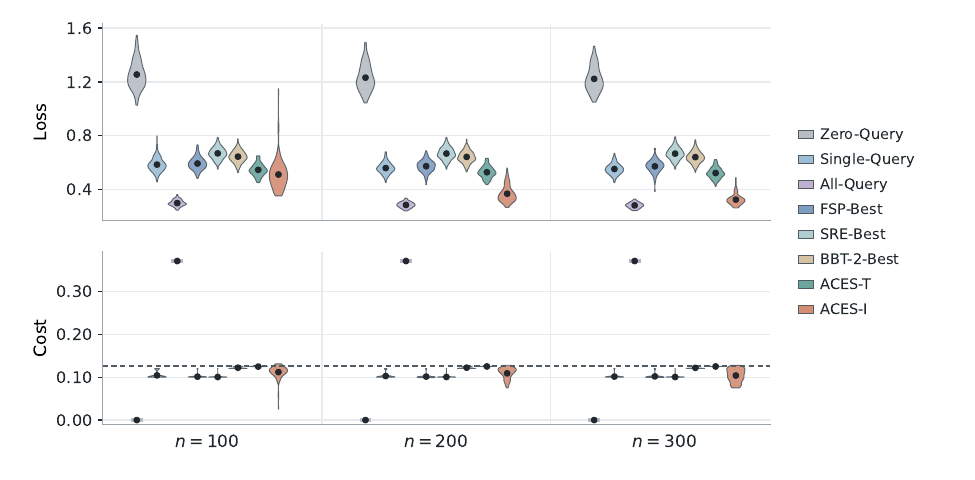}
	\caption{Test MSE (upper panel) and query cost (lower panel) in
		Scenario~2. Violins show distributions over 100 replications, dots
		mark means, horizontal bars mark quantities that are constant across
		replications, and the dashed line marks \(B\).}
	\label{fig:sim2-performance}
\end{figure}

\begin{figure}[htbp]
	\centering
	\includegraphics[width=0.85\textwidth]{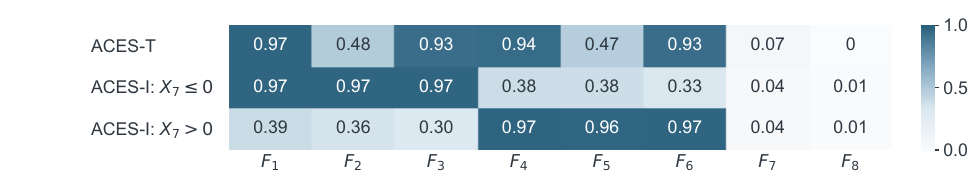}
	\caption{Fraction of test inputs on which ACES-T and ACES-I query each model
in Scenario~2 at \(n=300\), averaged over replications, with ACES-I shown
separately for \(X_7\leq0\) and \(X_7>0\).}
	\label{fig:sim2-calls-part}
\end{figure}

In Scenario~2, both ACES methods again have lower mean test MSE than all benchmarks except All-Query at every sample size
(Figure~\ref{fig:sim2-performance}). Moreover, ACES-I improves on ACES-T, lowering the mean test MSE at \(n=300\) from \(0.524\) to \(0.327\) and the mean query cost from \(0.125\) to \(0.104\). The query frequencies in Figure~\ref{fig:sim2-calls-part} explain this gain. Because ACES-T queries a fixed collection within each replication, its query
frequencies do not depend on \(X_7\), and the figure shows them in a single row. ACES-I instead queries each of models 1, 2 and 3 on about
\(97\%\) of inputs with \(X_7\leq0\) and each of models 4, 5 and 6 on \(96\%\)
to \(97\%\) of inputs with \(X_7>0\), and it queries each model of the other
collection on only \(30\%\) to \(39\%\) of inputs. Supplementary Figure~\ref{fig:sim2-calls-all} reports the query frequencies of all methods in Scenario~2.

\section{Laptop price prediction}\label{sec:real-data}
We apply the methods to the Amazon Laptop Product Catalog, in which
each observation is a laptop listing with structured attributes, a title
and description, a product image, and a price in US dollars
(\url{https://huggingface.co/datasets/naga-jay/amazon-laptop-product-catalog}).
Different external models can predict the price from different parts of a
listing, and we consider six. Tab-N is a tabular foundation model that uses the
numerical attributes. Jev-T and Qwen-T are text models that use the title and
description. GPT-I and Gemini-I are vision-language models that use the image,
and Gemini-J is a configuration of Gemini that receives all three modalities.
As in Remark~\ref{rem:external-library}, Gemini-I and Gemini-J use the same
foundation model with different inputs and count as two models. The response of
each model is converted into a price prediction, which enters the prediction
rules on the log scale. Each query is charged, and the estimated costs of the
six models in USD per 10000 predictions are 0.0054, 0.329, 0.537, 1.69, 4.08
and 4.93, in the order above. All fitting and budget selection use these costs,
and the reported costs are those realized on the test listings, which vary
around them. Supplementary~\ref{sec:laptop-implementation} lists the underlying
models in Table~\ref{tab:real-laptop-services} and describes how the costs are
estimated and how the responses are converted into price predictions.

After removing records with invalid prices, attributes, text, or images, 5614 of
the 30443 listings remain. We reserve 1000 of them for configuring the models
and estimating their costs and use the remaining 4614 for evaluation. The
outcome is the log listed price. Here \(\mathbf b_q(\bX)\)
contains an intercept and 46 features, namely 20 cubic B-spline features of
screen size, memory, storage, and weight, 17 text features given by
singular-value scores of TF-IDF weights and the log text length, and 9 image
summaries such as dimensions and color moments. The prediction rules are fitted by least squares on these features and the acquired outputs.

The analysis is repeated over 20 random splits of the 4614 listings into
training, validation, and test samples of sizes 2307, 1153, and 1154, with
listings that share an image or description assigned to the same sample.
Accuracy is measured by the test root mean squared error on the log scale
(log-RMSE). The methods are those of Section~\ref{sec:simulation}, and, as
there, Zero-Query, Single-Query, All-Query, and ACES-T are fitted on the pooled training and
validation samples because they require no validation step. ACES-I uses four
prespecified contrast classes and 44 cost multipliers, as detailed in
Supplementary~\ref{sec:laptop-implementation}. Since a fixed collection becomes
affordable only when the budget reaches its total cost, all methods are
evaluated at the 64 subset costs and at three fractions of the cheapest
model's cost. We focus on three budgets: \(B_1=0.334\) covers Tab-N and Jev-T,
\(B_2=4.08\) equals the cost of Gemini-I and lies just below that of Gemini-J,
and \(B_3=6.64\) covers the five single-modality models, which together with Gemini-J cost 11.6.

\begin{figure}[htbp]
\centering
\includegraphics[width=0.8\textwidth]{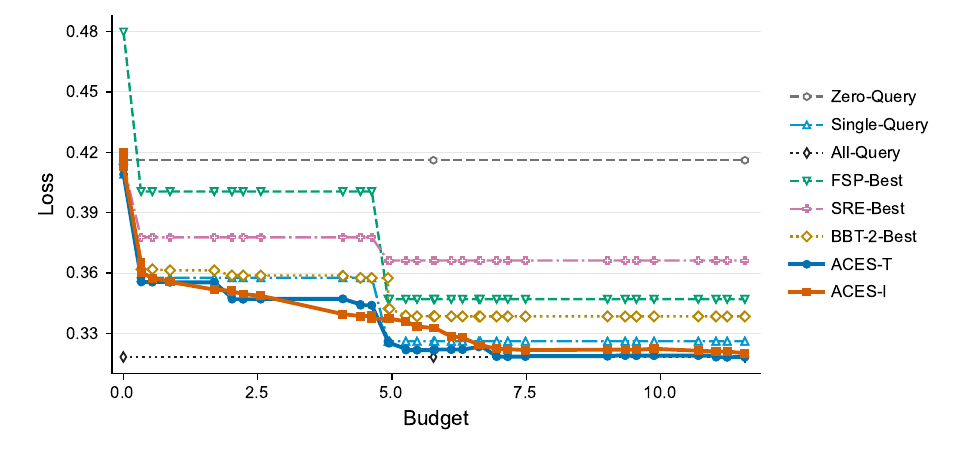}
\caption{Prediction loss (log-RMSE) against the budget at $n=2307$ for the eight methods. The budget is shown in USD per 10000 predictions.}
\label{fig:real-laptop-budget}
\end{figure}

\begin{figure}[htbp]
\centering
\includegraphics[width=\textwidth]{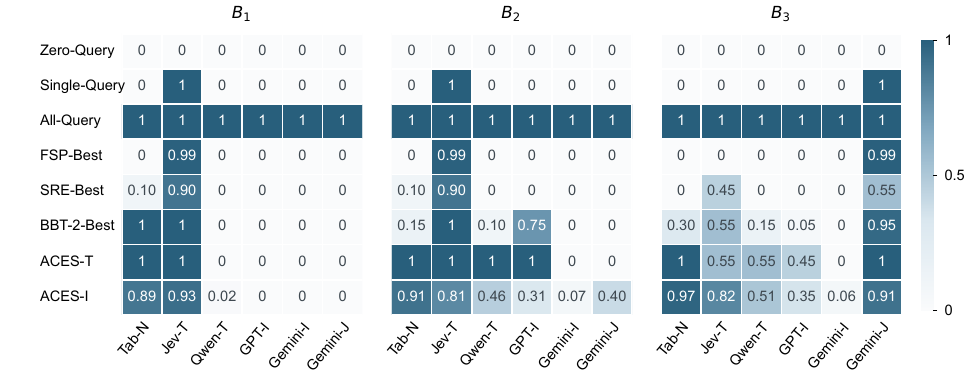}
\caption{Fraction of test listings on which each external model is queried
in laptop price prediction.}
\label{fig:real-laptop-queries}
\end{figure}

Figure~\ref{fig:real-laptop-budget} plots the mean test log-RMSE against the
budget for the eight methods at $n=2307$. At every budget at
which they are available, ACES-T and ACES-I have smaller error than FSP-Best,
SRE-Best, and BBT-2-Best, and ACES-T is never worse than Single-Query. Once Jev-T becomes affordable, Single-Query remains at a log-RMSE of 0.358 until Gemini-J becomes affordable, whereas ACES-T continues to improve as the budget admits additional outputs. At a budget of 6.95, about 60\% of the cost of All-Query, ACES-T attains a log-RMSE of 0.318, within 0.001 of All-Query, and larger budgets yield little further improvement. As in Scenario~1, the gain comes from combining complementary outputs, consistent with Theorem~\ref{thm:target-task-risk}. Table~\ref{tab:real-laptop-performance} and Figure~\ref{fig:real-laptop-queries} show that ACES-T remains more accurate even when it queries the same models as a benchmark. For
example, ACES-T and BBT-2-Best both query Tab-N and Jev-T for every
listing at \(B_1\). ACES-T attains a lower log-RMSE, 0.355 compared with
0.362, even though BBT-2-Best is selected using test data. The two
methods incur the same realized cost, 0.345, which slightly exceeds
\(B_1\) because the budget constraint is imposed on estimated costs.

\begin{table}[htbp]
\centering
\caption{Accuracy and realized cost in laptop price prediction at
\(n=2307\). Entries are means over 20 splits, with standard deviations in
parentheses. Cost is the realized cost in USD per 10000 predictions.
All-Query exceeds all three budgets.}
\label{tab:real-laptop-performance}
\renewcommand{\baselinestretch}{1}\small
\setlength{\tabcolsep}{2pt}\renewcommand{\arraystretch}{1.15}
\begin{tabular}{l cc cc cc}
\toprule
& \multicolumn{2}{c}{$B_1$} & \multicolumn{2}{c}{$B_2$}
& \multicolumn{2}{c}{$B_3$}\\
\cmidrule(lr){2-3}\cmidrule(lr){4-5}\cmidrule(lr){6-7}
Method & Log-RMSE & Cost & Log-RMSE & Cost & Log-RMSE & Cost\\
\midrule
Zero-Query    & 0.416 (0.028) & 0 & 0.416 (0.028) & 0 & 0.416 (0.028) & 0\\
Single-Query  & 0.358 (0.026) & 0.340 (0.018) & 0.358 (0.026) & 0.340 (0.018)
           & 0.326 (0.016) & 5.000 (0.130)\\
FSP-Best   & 0.400 (0.023) & 0.335 (0.017) & 0.400 (0.023) & 0.335 (0.017)
           & 0.347 (0.022) & 4.960 (0.130)\\
SRE-Best   & 0.378 (0.032) & 0.310 (0.105) & 0.378 (0.032) & 0.310 (0.105)
           & 0.366 (0.020) & 2.880 (2.350)\\
BBT-2-Best & 0.362 (0.021) & 0.345 (0.018) & 0.359 (0.020) & 1.640 (0.640)
           & 0.338 (0.017) & 5.100 (1.180)\\
ACES-T     & 0.355 (0.024) & 0.345 (0.018) & 0.347 (0.019) & 2.570 (0.090)
           & 0.324 (0.020) & 6.240 (0.500)\\
ACES-I     & 0.360 (0.022) & 0.333 (0.026) & 0.340 (0.019) & 3.300 (0.790)
           & 0.324 (0.018) & 5.970 (0.960)\\
\midrule
All-Query  & \multicolumn{6}{c}{Log-RMSE 0.318 (0.017) at cost 11.600 (0.230)}\\
\bottomrule
\end{tabular}
\end{table}

Compared with ACES-T, ACES-I can query an expensive model only for the
listings on which it is worth its cost. This matters when the strongest
model is too expensive to query for every listing. At
intermediate budgets such as \(B_2\), a single Gemini-J query costs more
than the budget, so no affordable fixed collection can include it, and
ACES-T queries the four least expensive models for every listing.
ACES-I queries fewer models on average but uses the average budget to
query Gemini-J for about 40\% of listings, as Figure~\ref{fig:real-laptop-queries} shows. It therefore spends 81\% of the budget and attains a log-RMSE of 0.340, compared with 63\% and 0.347 for ACES-T. Querying a model for only part of the listings also
lets ACES-I improve on Zero-Query at budgets below the cost of the
cheapest model. When Gemini-J first becomes affordable at \(B=4.93\), Single-Query and ACES-T query it for every listing and attain a log-RMSE of 0.326. ACES-I has a higher error of 0.337 at a lower realized cost of 3.88 versus 5.00. A fixed collection then already contains the strongest model, so choosing models input by input adds estimation error without improving which models are queried. This gap narrows as the budget grows: Table~\ref{tab:real-laptop-performance} shows that at \(B_3\) both ACES methods attain a log-RMSE of 0.324 and that ACES-I still costs less, 5.97 versus 6.24.

\section{Discussion}\label{sec:conclusion}

We have treated the outputs of paid foundation models as inputs to a prediction rule. The task's own data then serve to measure how much each output adds
beyond the features and the other outputs and, under the budget, to decide
which models to query and how to combine their outputs with the features. Because both ACES methods retain Zero-Query as a candidate, a
model that adds nothing costs only a small estimation error.

The framework complements fine-tuning \citep{hu2022lora}
and in-context learning \citep{brown2020language}.
Fine-tuning requires access to the model parameters, which most commercial
models keep private, and even its parameter-efficient versions need
substantial task data and computing resources when the model is large. In-context
learning, in turn, can use only as many examples as fit in the prompt,
and every additional example raises the cost of each query. Our methods need only the outputs of the models on the training sample, and
they use every observation in that sample to fit low-dimensional prediction rules. The approaches can also
be combined, because an adapted model is simply one more paid output that the
framework can query and combine with the others. In the laptop application,
for instance, Tab-N uses 1000 listings with known prices as its context. The paid outputs also need not come from foundation models, since the framework applies
whenever predictions of the outcome can be purchased input by input from
sources of differing cost and accuracy, such as expert assessments or
simulators of different fidelity.

Several questions remain open. First, our analysis requires every external output to be recorded for every observation. When
the data were collected under an earlier acquisition rule, only some outputs
are recorded for each observation, and whether an output is recorded may
depend on those already acquired. In that case, doubly robust policy learning
under such informative missingness \citep{zhou2026costoptimal} may offer a
starting point. Second, our prediction classes are linear in their
coefficients. Basis expansions of the outputs preserve this structure, as noted in Section~\ref{sec:task-method}, but fully nonlinear learners, such as
neural networks or nonlinear aggregation within each collection
\citep{wang2024fusing}, can be fitted only approximately, so the
branch-and-bound search may lose its exactness and the risk bounds would
require complexity measures other than the dimension. Third, as the number of
available models grows, deciding which outputs to buy and how to combine them
becomes a variable selection problem in which the candidate variables are
paid, possibly biased and strongly correlated predictions of the outcome. In
the risk bound of Theorem~\ref{thm:target-task-risk}, the price of selecting
among the \(N_B\) affordable collections is of order \((\log N_B)/n\), and
\(\log N_B\leq K_B\log(eK/K_B)\), where \(K_B\geq1\) is the size of the
largest affordable collection. As in best-subset selection, this price
therefore grows only logarithmically in \(K\) when \(K_B\) is fixed.
Computation is the harder obstacle, because the backward recursion of ACES-I
runs over all \(2^K\) collections, and the number of collections examined by
the branch-and-bound search can grow exponentially when \(K_B\) grows with \(K\). Restricting the query
rules reduces this cost. If the models are queried in a fixed order, such as
increasing cost as in cascades \citep{chen2024frugalgpt}, ACES-I only decides
when to stop, and its recursion runs over \(K+1\) collections, at the price of
a smaller class of query rules. Computationally efficient procedures for many models, together with the
choice of which models to make available in the first place, are left for future work.

\medskip
\noindent\textbf{Funding.} This work received no specific funding.
\textbf{Disclosure statement.} The authors report no competing interests.
\textbf{Use of generative AI.} The authors used OpenAI Codex (GPT-6 Astra) to
edit the language and to check proofs, code and numerical results. The authors
checked the tool's terms of use, confirmed the originality and accuracy of all
content, and take full responsibility for its integrity, including the accuracy
of the references.

{\spacingset{1}\footnotesize
\bibliographystyle{chicago-review}
\bibliography{foundation_model_assisted_learning_clean_refs}
}
\end{document}